\documentclass[conference]{IEEEtran}
\IEEEoverridecommandlockouts
\usepackage{cite}
\usepackage{algorithmic}
\usepackage{graphicx}
\usepackage{textcomp}
\usepackage{xcolor}
\usepackage{epsfig}
\usepackage{amsfonts}
\usepackage{amssymb,amsmath,multicol}
    
\newtheorem{remark}{Remark}
    \newcommand{\REMARK}[1]{ \begin{remark} #1 \end{remark}} 

\usepackage{listings}

\usepackage{cancel}
    
\newcommand{\COMMENT}[1]{}

\usepackage{tabularx}
\usepackage{enumitem}
\usepackage{float}
\usepackage{subcaption}
\usepackage{array}
\usepackage{ragged2e}
\usepackage[table]{xcolor}
\usepackage{booktabs}
\usepackage[hidelinks]{hyperref}
\usepackage[normalem]{ulem}

 \makeatletter
    \newcommand{\linebreakand}{%
      \end{@IEEEauthorhalign}
      \hfill\mbox{}\par
      \mbox{}\hfill\begin{@IEEEauthorhalign}
    }
    \makeatother

\def\BibTeX{{\rm B\kern-.05em{\sc i\kern-.025em b}\kern-.08em
    T\kern-.1667em\lower.7ex\hbox{E}\kern-.125emX}}

\begin{document}

\title{Estimation of Ground Reaction Forces from Kinematic Data during Locomotion}

\author{ 
    \IEEEauthorblockN{ 
        Gautami Golani\textsuperscript{1,\ensuremath{\dag}}, 
        Dong Anh Khoa To\textsuperscript{2,\ensuremath{\dag}}, 
        Ananda Sidarta\textsuperscript{3}, 
        Arun-Kumar Kaliya-Perumal\textsuperscript{3},\\ 
        Oliver Roberts\textsuperscript{3}, 
        Lek Syn Lim\textsuperscript{3}, 
        Jim Patton\textsuperscript{4}, 
        Domenico Campolo\textsuperscript{1,*}
    } 
    \IEEEauthorblockA{\textsuperscript{1}School of Mechanical and Aerospace Engineering, Nanyang Technological University, Singapore}
    
    \IEEEauthorblockA{\textsuperscript{2}College of Engineering and Computer Science, VinUniversity, Vietnam} 
    \IEEEauthorblockA{\textsuperscript{3}Rehabilitation Research Institute of Singapore, Nanyang Technological University, Singapore} 
    \IEEEauthorblockA{\textsuperscript{4}Biomedical Engineering, University of Illinois Chicago, USA} 
     
    \thanks{\textsuperscript{\ensuremath{\dag}}These authors contributed equally to this work.} 
    \thanks{\textsuperscript{*}Corresponding author (Email: d.campolo@ntu.edu.sg)}
    \thanks{Funding support provided by Rehabilitation Research Institute of Singapore (Grant ID: 021099-00001), a tripartite collaboration between the Nanyang Technological University (NTU), the Agency for Science, Technology and Research (A*STAR), and the National Healthcare Group (NHG Health).}
}

\maketitle

\begin{abstract}
Ground reaction forces (GRFs) provide fundamental insight into human gait mechanics and are widely used to assess joint loading, limb symmetry, balance control, and motor function. Despite their clinical relevance, the use of GRF remains underutilised in clinical workflows due to the practical limitations of force plate systems. In this work, we present a force-plate-free approach for estimating GRFs using only marker-based motion capture data. This kinematics-only method estimates and decomposes GRFs without force plates, making it well suited for scalable clinical deployment. Using kinematic data from sixteen body segments, we estimate the centre of mass (CoM) and compute GRFs, which are subsequently decomposed into bilateral components through a minimization-based approach. Through this framework, we can identify gait stance phases and provide access to clinically meaningful kinetic measures without a dedicated force plate system. Experimental results demonstrate the viability of CoM and GRF estimation based solely on kinematic data, supporting force-plate-free gait analysis.
\end{abstract}


\section{Introduction}
Ground reaction forces are one of the most fundamental measures in human movement analysis. Defined as the forces exerted by the ground on the body during standing, walking, or performing functional tasks, they provide direct insight into how individuals load, propel, and stabilize themselves \cite{McLester2008}. Although force plates and kinetic gait analysis have long been used in research laboratories, the clinical relevance of GRF has grown substantially with advances in rehabilitation technology, motion analysis systems, and instrumented assessment tools.

\begin{table*}[t]
\centering
\caption{Clinical interpretation of ground reaction force (GRF) components}
\label{table:clinical_interpretation}
\begin{tabularx}{\textwidth}{>{\RaggedRight\arraybackslash}m{4cm} m{4cm} m{9cm}}
\hline
\textbf{GRF Component} & \textbf{What It Reflects} & \textbf{Clinical Examples} \\
\hline

\textbf{Vertical GRF (vGRF)} &
Weight acceptance, impact loading, loading rate, and stance-phase support & \vspace{0.5em}
\begin{itemize}[leftmargin=*, nosep]
    \item \textbf{Knee osteoarthritis:} reduced first peak vGRF and loading rate indicate pain-related unloading and correlate with disease severity \cite{Bjornsen2023}.
    \item \textbf{Frailty/ageing:} lower vGRF during gait and sit-to-stand reflects diminished strength and predicts mobility decline \cite{Hirano2022}.
    \item \textbf{ACL injury/reconstruction:} reduced vGRF on the operated limb during landing signals persistent asymmetry and increased re-injury risk \cite{Ahmadi2022}.
\end{itemize}
\\
\hline

\textbf{Medio-lateral GRF (ML-GRF)} &
Lateral stability, balance control, weight-shifting ability & \vspace{0.5em}
\begin{itemize}[leftmargin=*, nosep]
    \item \textbf{Stroke:} increased ML-GRF variability and asymmetry indicate impaired lateral stability and higher fall risk \cite{Osada2022}. 
    \item \textbf{Parkinson's disease:} elevated temporal fluctuation of ML-GRF during stance reflect postural instability \cite{Minamisawa2012}.
\end{itemize}
\\
\hline

\textbf{Antero-posterior GRF (AP-GRF)} &
Control of forward momentum, braking forces, push-off (propulsion) & \vspace{0.5em}
\begin{itemize}[leftmargin=*, nosep]
    \item \textbf{Stroke:} reduced propulsive GRF on the paretic side indicates plantarflexor weakness and contributes to slow gait \cite{Bowden2006}.
    \item \textbf{Sarcopenia/weakness:} diminished push-off forces during walking reflect reduced plantarflexor power and mobility limitations \cite{Franz2016}.
    \item \textbf{Post-total knee replacement:} decreased push-off represents an attempt to minimize compressive force on the joint \cite{Mandeville2007}.
\end{itemize}
\\
\hline
\end{tabularx}
\end{table*}

In clinical practice, GRF offers a unique window into joint loading, limb symmetry, postural stability, and motor control, making it a valuable indicator across a wide range of musculoskeletal and neurological conditions \cite{Liu2024, Liu2023}. Alterations in GRF patterns can reflect disease severity, compensatory strategies, pain avoidance, muscle weakness, or impaired neuromuscular coordination \cite{Schneider1983, Giakas1997}. Table \ref{table:clinical_interpretation} provides a clinical interpretation of the vertical, medio-lateral, and antero-posterior GRF components.

Taken together, GRF represents an underutilised yet highly informative clinical tool. A clearer understanding of GRF applications across clinical conditions may support broader use in functional assessment, rehabilitation planning, and longitudinal patient monitoring. Despite these advantages, the integration of GRF into routine clinical workflows remains limited. Traditional force plate systems require space, cost, and technical expertise, and results are often confined to specialised gait laboratories. However, recent advances in wearable sensors, instrumented walkways, and markerless computer vision systems are making kinetic assessment more accessible, portable, and scalable \cite{Liu2024, Yang2025, Ren2008}. These developments create opportunities to incorporate GRF-based measures into clinical decision-making, risk stratification, and longitudinal monitoring across diverse settings.

Several methods estimate GRFs from kinematic data, broadly categorized into physics-based and data-driven approaches. Physics-based methods calculate total GRF by applying Newton-Euler equations to whole-body segmental accelerations derived from marker-based \cite{Patoz2021} or inertial measurement unit (IMU)-based \cite{Ancillao2018} motion capture. However, decomposing this total GRF into bilateral components during the statically indeterminate double-stance phase remains challenging. Proposed analytical solutions, such as transfer or membership functions \cite{Ancillao2018}, are largely limited to normal walking analysis. Parametric curve-fitting models achieve low errors but rely on known single-stance boundaries, assume stereotyped GRF profiles, and are validated only on healthy adults \cite{Samadi2017}. Musculoskeletal optimization using artificial contact actuators \cite{Fluit2014} offers another approach, but incurs a high computational load and relies on simplified knee and foot joint models. Alternatively, data-driven machine learning approaches, including neural networks \cite{Ancillao2018} and predictive models applied to markerless kinematics \cite{Lichtwark2024}, circumvent these biomechanical assumptions but remain constrained by their reliance on large training datasets.

\begin{figure}[h!]
    \centering
    \includegraphics[width=0.8\linewidth]{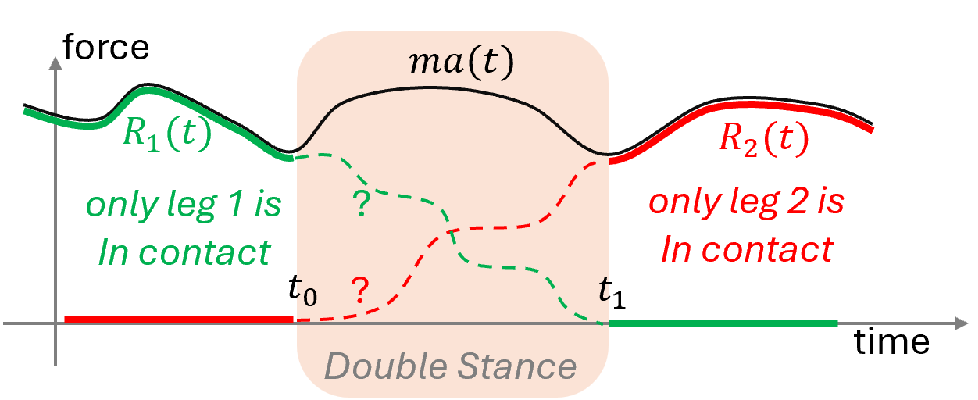}
    \caption{During the \textit{Double-Stance} phase of typical locomotion pattern, an ambiguity arises as to the individual contribution of GRF.}
    \label{fig:min_GRF}
\end{figure}

In this work, we present a force-plate-free approach to gait analysis through which we estimate the CoM and compute the estimated GRF using whole-body kinematic data from sixteen body segments. In addition to this, we also decompose the GRF into its individual components, as shown in Fig. \ref{fig:min_GRF}. GRFs are computed directly from segment accelerations and then decomposed into bilateral components using a minimization-based technique, eliminating the need for specialized force plate measurements while remaining computationally light. This enables identification of gait stance phases and provides a kinematics-based framework for analyzing human locomotion.

\section{Methods} \label{Methods}
\subsection{Experimental Protocol} 
The current work draws from an earlier 10-m walk test dataset comprising approximately 500 participants of Asian descent \cite{Liang2020}. It represents a proof-of-concept analysis of four representative participants aimed at demonstrating the proposed framework and opening avenues for future investigation in larger cohorts. Briefly, the experiment involved a marker-based motion capture system (Miqus M3 Mocap, Qualisys AB, Sweden) and force plates (Kistler, Switzerland) to extract kinematic and kinetic information, respectively. Healthy participants aged 21-80 years, with no prior neurological and musculoskeletal conditions were recruited for the study (see Table \ref{table:participant_deomographics}). Retroreflective markers with a diameter of 12.5 mm were placed at anatomical landmarks according to a modified Calibrated Anatomical System Technique (CAST) protocol. The participant was asked to walk straight for a total distance of 10 meters indicated by a demarcated start line at one end and another stop line at the other end. Participants performed the task in both directions. In each trial, the participant was required to have a clean heel strike on each force plate, i.e. only one foot entirely stepped on one plate. In this way, the ground reaction force from the individual foot can be assessed separately. The trials would end after each participant achieved three clean steps with each foot.

\begin{table}[H]
\caption{Participant Demographics}
 \label{table:participant_deomographics}
\rowcolors{2}{gray!10}{white}
\centering
\begin{tabularx}{\linewidth}{X X X X X}
\toprule
\rowcolor{gray!20}
\textbf{Participant} & \textbf{Age} & \textbf{Gender} & \textbf{Height (cm)} & \textbf{Mass (kg)} \\
\midrule
P1 & 72 & M & 174.5 & 83.1 \\
P2 & 74 & F & 151 & 51.3 \\
P3 & 72 & M & 159 & 63 \\
P4 & 21 & M & 166.5 & 66.2 \\
\bottomrule
\end{tabularx}
\end{table}

Data acquisition for both motion capture and force plates was controlled using the Qualisys Track Manager (QTM) software. Marker trajectories were recorded at 200Hz, with analogue and force signals at 2000 Hz, respectively.

\subsection{Centre of Mass (CoM) Estimation}
The whole-body Centre of Mass (CoM) was estimated using a weighted segmental analysis based on a sixteen-segment rigid body model. The body was segmented into the Head \& Neck, Thorax, Abdomen, Pelvis, and bilateral Upper Arms, Forearms, Hands, Thighs, Shanks, and Feet according to Dumas et al. model \cite{Dumas2017}. In some motion capture experiments focusing on lower-limb tasks such as gait, hand markers may be intentionally omitted to reduce marker placement complexity and streamline data processing. As a consequence, the full marker set required by the Dumas segment model is not available, preventing the reconstruction of the hand local coordinate system and centre of mass. To address this limitation, the hand CoM was instead estimated using an anthropometry-based approach following Winter et al. \cite{Winter2009}, where the hand is modeled as a rigid segment whose length is defined as a fixed proportion of the forearm length. Specifically, hand length was assumed to be 74$\%$ of the forearm length, and the hand CoM was located at the midpoint of this segment along the forearm–hand axis originating from the wrist joint centre. This approach enables consistent whole-body CoM estimation despite reduced upper-limb marker availability.

Motion capture data were recorded using the marker placement protocol described in \cite{Liang2020}. For each segment, local coordinate systems were constructed to define the position and orientation of each section in the global space. To accommodate the specific marker configurations available in this study, the segment reference frames were defined using a modification of the International Society of Biomechanics (ISB) standard mentioned in \cite{Dumas2017}. Specifically, the local X-axis was aligned with the antero-posterior direction, the Y-axis with the medio–lateral direction, and the Z-axis with the vertical direction. Segment inertial parameters, including mass and local CoM location, were computed using gender-specific regression equations and anthropometric tables adapted from \cite{Dumas2017}. The mass of each segment ($m_i$) was calculated as a percentage of the subject's known total body mass ($M_{subject}$), such that 
\begin{equation}
    m_i = M_{subject} \cdot P_{mass, i}
\end{equation}
where $P_{mass, i}$ represents the segment-specific mass ratio. Joint centres, particularly the cervical, thoracic, lumbar, and hip joints, were estimated mathematically from surface markers using the geometric deviation percentage provided by the Dumas's regressions \cite{Dumas2017}. The local Centre of Mass ($CoM_i$) for each segment was determined relative to a reference joint centre ($\text{Origin}_i$) using geometric offset proportions ($P_{AP}, P_{ML}, P_{SI}$) along the segment's orthonormal basis vectors ($\vec{u}_x, \vec{u}_y, \vec{u}_z$):
\begin{equation}
\vec{r}_{CoM, i} = \text{Origin}_i + L_i \left( P_{AP, i}\vec{u}_x + P_{ML, i}\vec{u}_y + P_{SI, i}\vec{u}_z \right)
\end{equation}

Where $L_i$ represents the segment length derived from the distance between the proximal and distal joint centres, calculated by taking the Euclidean coordinate norm of two end points. Finally, the instantaneous position of the total CoM ($\vec{r}_{CoM}$) was computed as the weighted average of the sixteen individual segment CoMs:

\begin{equation}
\vec{r}_{CoM}(t) = \frac{\sum_{i=1}^{16} m_i \vec{r}_{i}(t)}{\sum_{i=1}^{16} m_i}
\end{equation}

where $\vec{r}_{i}(t)$ denotes the position vector of the $i$-th segment's centre of mass at time $t$. This comprehensive multi-segment trunk model (separating thorax, abdomen, and pelvis) was selected to provide higher bio-fidelity in tracking mass distribution during dynamic movement compared to single-segment trunk models. By tracking the 3D spatial coordinates of 16 individual body segments frame-by-frame, the algorithm calculates the continuous trajectory of the subject's whole-body centre of mass over time.

\subsection{Event Detection}
\begin{figure}[h]
    \centering
    \begin{subfigure}{\linewidth}
        \centering
        \includegraphics[width=0.8\columnwidth]{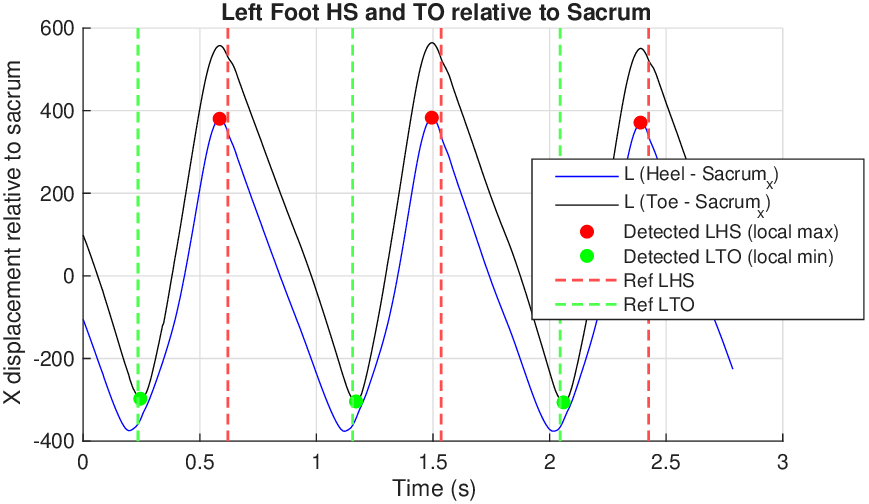}
        \label{fig:left_foot}
    \end{subfigure}
    \medskip
    \begin{subfigure}{\linewidth}
        \centering
        \includegraphics[width=0.9\linewidth]{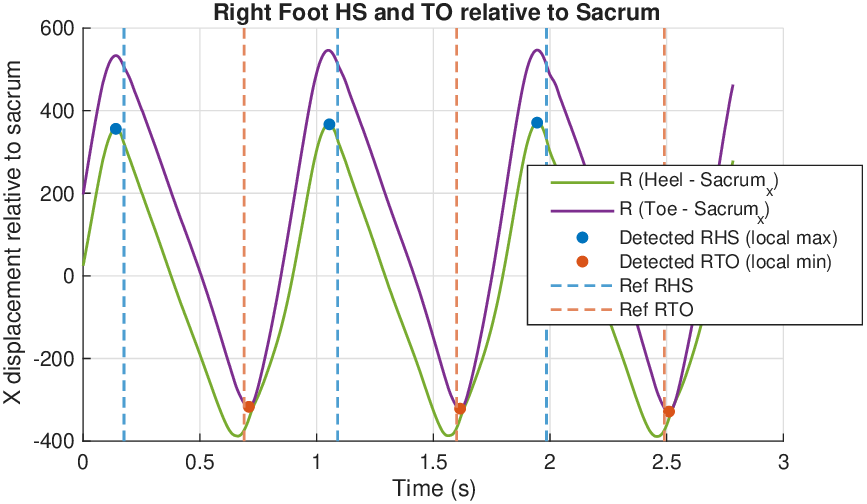}
        \label{fig:right_foot}
    \end{subfigure}
    \caption{Gait events detected using the sacrum-based kinematic method proposed by Zeni et al. \cite{Zeni2008} and validated against Visual3D results. HS and TO events for the left (top) and right (bottom) feet were identified from foot kinematics relative to the sacrum during gait.}
    \label{fig:event_detection}
\end{figure}

Gait analysis was performed by segmenting the gait cycle into stance and swing phases based on the identification of key gait events, namely heel strike (HS) and toe-off (TO). In this study, gait events were detected using both GRF profiles and a kinematics-based approach proposed by Zeni et al. \cite{Zeni2008}, wherein HS and TO are identified from extrema in the antero–posterior displacement of the heel and toe markers relative to the sacrum (see Fig. \ref{fig:event_detection}).
\subsection{Method to Decompose Ground Reaction Force}
Consider a `two-legged` body (not necessarily rigid) of known \textit{total} mass $m$, subjected to gravity as well as to two forces originating from leg-ground interaction, i.e. GRF. In Cartesian\footnote{This assumption is essential to be able to write Newton's law.} coordinates, we can evaluate Newton's second law at the centre-of-mass and project it along any axis, for example the vertical one, providing the scalar constraint:
\begin{equation}\label{eq:Newton}
    R_1(t) + R_2(t) = ma(t)
\end{equation}
where $R_1$ and $R_2$ represent the Ground Reaction Forces, respectively, for the first and second leg.

Assume that the acceleration $a(t)$ is known\footnote{In the vertical direction, one would need to \textit{subtract} gravity, i.e. $a=\ddot z+g$, assuming that the $z$-axis pointing upwards.}, e.g. as measured via some motion tracking system. 

As sketched in Fig.~\ref{fig:min_GRF}, typical locomotion can be segmented into Single-Stance (SS) and Double-Stance (DS) phases.
During SS, i.e. when only one of the legs is in contact with the ground, an additional condition would be that either $R_1(t)=0$ or $R_2(t)=0$, so there would be no ambiguity in determining the non-zero GRF from Eq. \eqref{eq:Newton}. 
However, in DS situations, Eq. \eqref{eq:Newton} alone is insufficient to fully determine the GRFs.

A possible way to resolve this ambiguity is to impose a minimization principle. A classical choice is the minimum torque-changed model \cite{Uno1989} which, in this case, leads to simple, closed-form solutions.

The full mathematical formulation and derivation are reported in Appendix~\ref{grf_decomposition}.

\section{Results and Discussion} \label{Results and Discussion}
\subsection{Centre of Mass Estimation}
\begin{figure}[h]
    \centering
    \begin{subfigure}[b]{0.48\textwidth}
        \centering
        \includegraphics[width=\linewidth]{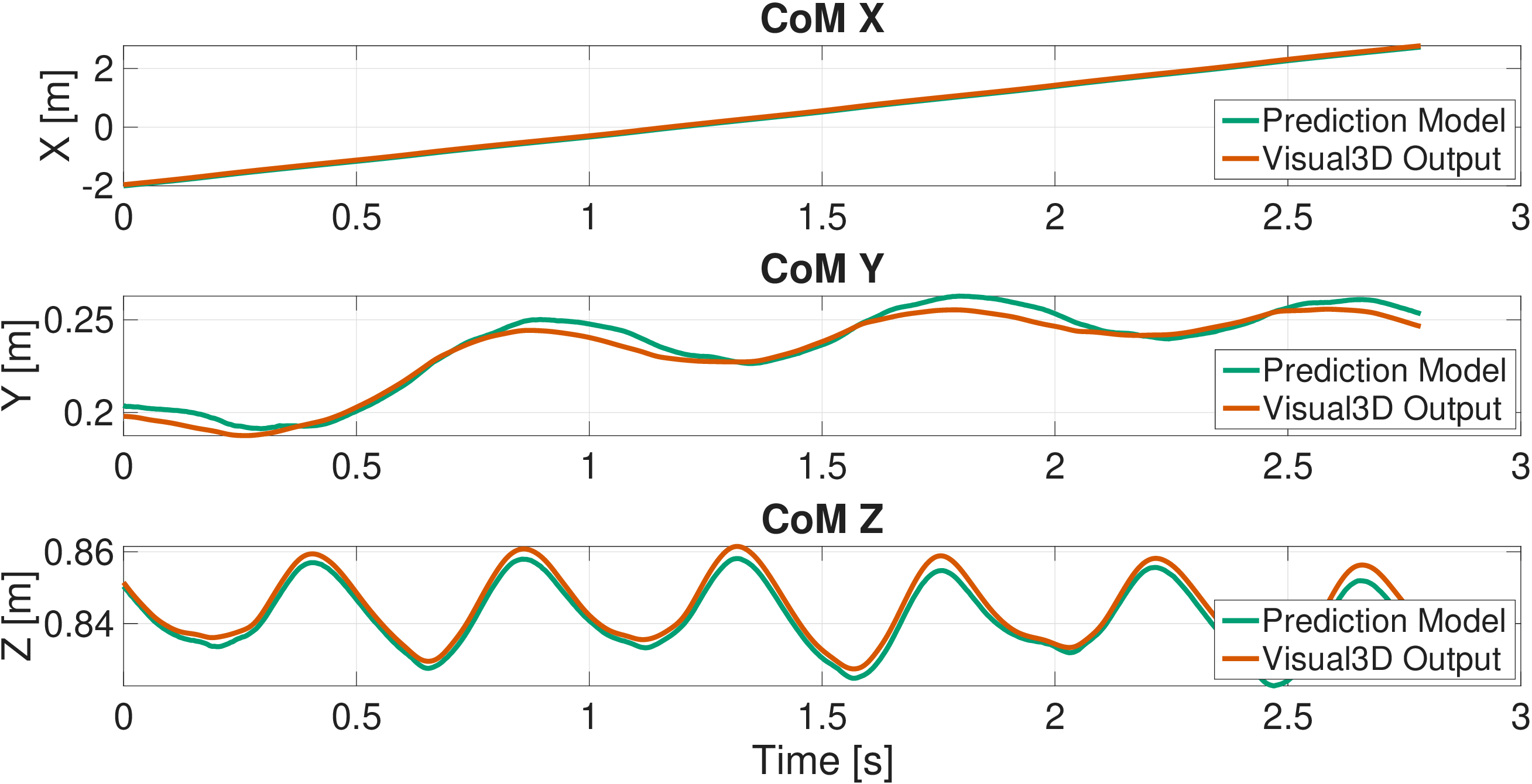}
        \caption{Visualization of CoM trajectories derived from the prediction model and Visual3D software}
        \label{fig:com_traj_1}
    \end{subfigure}
    \hfill 
    \begin{subfigure}[b]{0.48\textwidth}
        \centering
        \includegraphics[width=\linewidth]{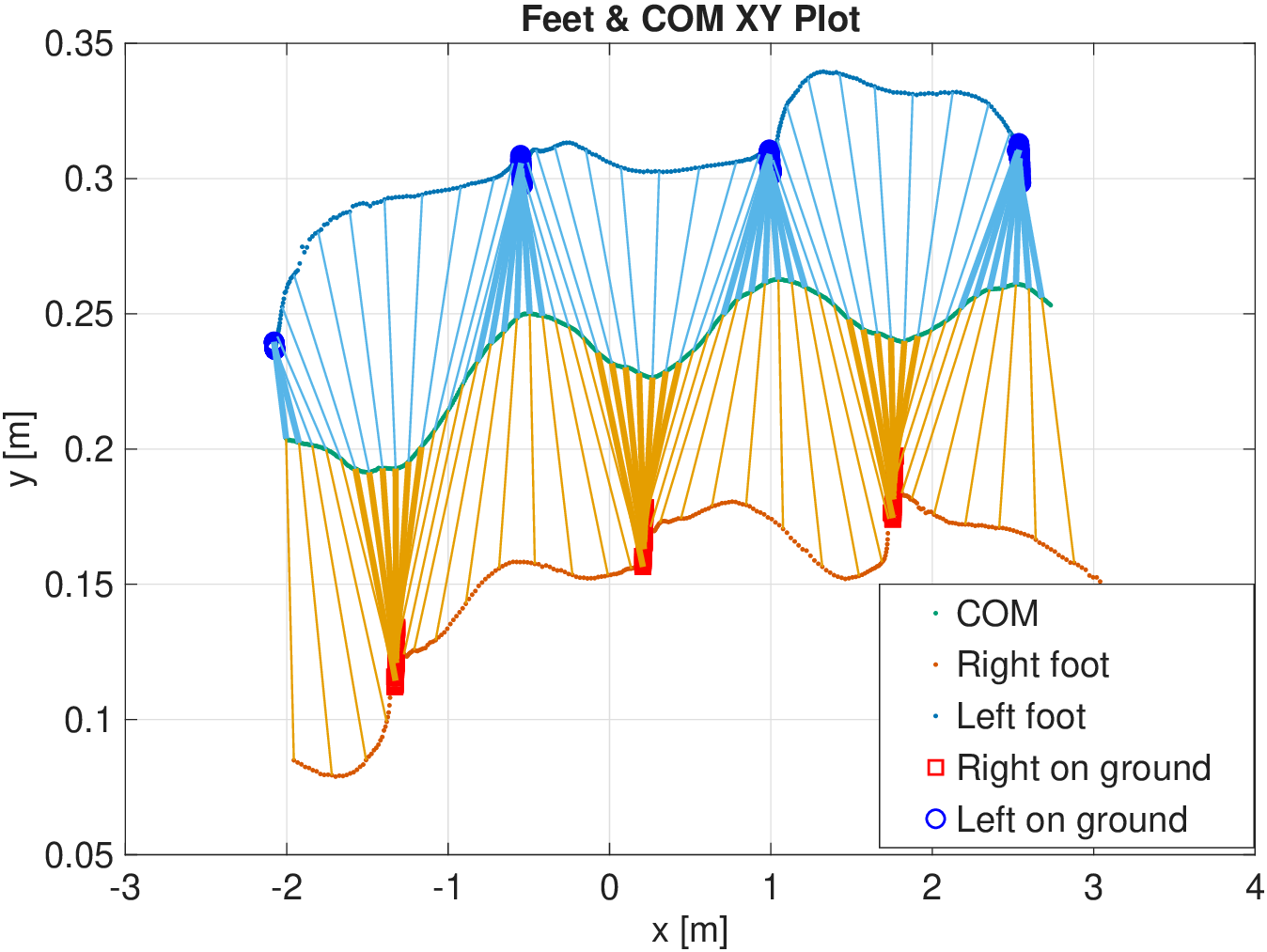}
        \caption{Trajectory of whole-body CoM and CoMs of two feet}
        \label{fig:com_traj_2}
    \end{subfigure}
    \caption{a) Comparison of estimated CoM trajectories between the custom sixteen segment model (Solid Line) and Visual3D benchmark (Dashed Line) for one representative participant. b) Planar trajectories of the centre of mass and both feet during walking, with stance phases identified from vertical foot position.}
    \label{fig:com_comparison}
\end{figure}

To evaluate the reliability of the proposed segmental analysis algorithm, the estimated CoM trajectories were validated against the industry-standard Visual3D software (C-Motion, Inc., USA) using identical marker inputs. The custom sixteen-segment model showed strong temporal agreement with the benchmark across the gait cycle, preserving the waveform morphology in all three planes (Fig. \ref{fig:com_traj_1}). Quantitatively, the method was most precise in the medio–lateral (Y) axis, with a Root Mean Square Error (RMSE) of only $0.31$–$0.52$ cm. In the antero–posterior (X) direction, the range RMSE was $3.93$–$6.19$ cm, which is still relatively accurate given that this axis contains the largest progression and therefore the greatest absolute variation. Notably, a systematic vertical offset was observed. The model estimate remained consistently $4.3$–$8.9$ cm higher than the Visual3D output. This deviation likely stems from differences in anthropometric assumptions. Our implementation uses the updated dataset of Dumas et al. \cite{Dumas2017}, whereas Visual3D relies on the Dempster \cite{Dempster1955} and Hanavan \cite{Hanavan1964} models. The newer model covered the increased bio-fidelity of the multi-segment trunk representation, which can capture subtle mass shifts that simplified rigid-body models may smooth out. After compensating for this constant bias by shifting the Z-prediction downward by $7.2$ cm, the remaining vertical (Z) difference fluctuated only within $0.01$–$2.93$ cm. The cumulative mean 3D RMSE across all trials is 5.09cm. Overall, the low error magnitudes support the algorithm’s accuracy for dynamic motion tracking. As this comparison is between model and model, it is difficult to draw conclusions about absolute accuracy. Therefore, a benchmark against force plate data would be more reliable. It is worth noting that any constant shifts in the CoM do not affect the prediction of GRFs, as accelerations are unaffected.


By monitoring whole-body CoM, simultaneously with bilateral CoMs of two feet, the participant's gait symmetry could be observed, (see Fig. \ref{fig:com_traj_2}). At each sampled time step, straight line segments connect the instantaneous CoM position to the corresponding positions of the left and right feet. Foot–ground contact is detected using the vertical (Z) coordinate of each foot. When the height of the foot falls below a predefined threshold $(Z<0.06m)$, the foot is considered to be in the stance phase. These stance events are explicitly marked on the XY projection using distinct symbols: red squares indicate right-foot contact, and blue circles indicate left-foot contact. During stance, the corresponding CoM-to-foot connection lines are drawn with increased thickness, highlighting moments when the body is mainly supported by that foot.

\subsection{Ground Reaction Force Estimation and Validation}
\begin{figure}[h!]
    \centering
    \includegraphics[width=1\columnwidth]{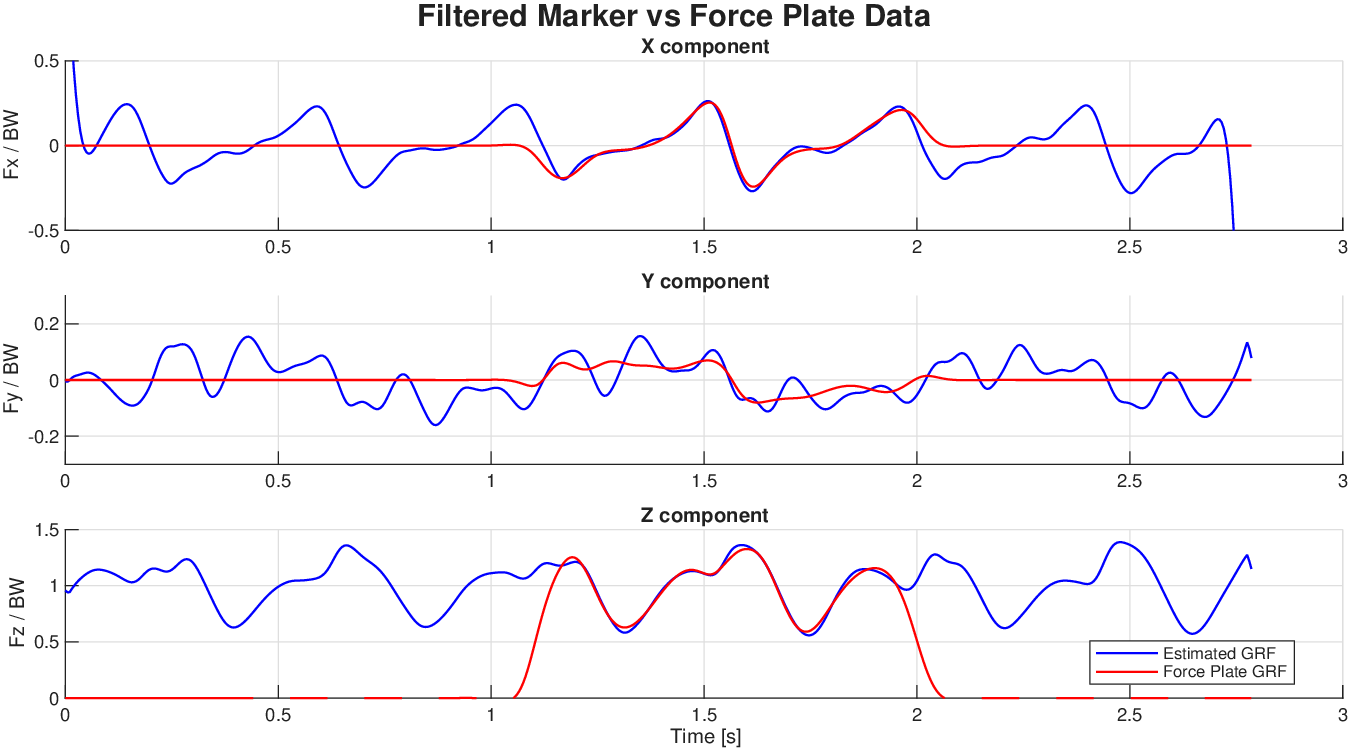}
    \caption{Filtered marker-based and force-plate GRFs for a representative participant. The vertical GRF during stance exhibits the characteristic double-peaked M-shaped profile, with peaks corresponding to loading response and push-off, and the intermediate region representing mid-stance and pre-swing phases.}
    \label{fig:filtered_marker_vs_force_data}
\end{figure}

GRFs acting on the foot were measured using force plates, providing a complete three-dimensional representation of the interaction between the foot and the ground. The force plates record the AP, ML and vertical components \cite{Winter2009}. These measurements were used to validate the GRFs estimated from marker-based kinematic data (Eq. \ref{eq:Newton}). To reduce high-frequency noise, both the marker and force plate data were filtered using the same Butterworth 4th-order low-pass filter with a standard 8 Hz cutoff frequency commonly used in gait analysis, thereby minimizing phase distortion between the two signals \cite{Kristianslund2012}. As seen in Fig. \ref{fig:filtered_marker_vs_force_data}, a strong agreement is observed between the two modalities.



\subsection{Bilateral Ground Reaction Force}
\begin{figure*}[!t]
    \centering
    \subfloat[Comparison between the predicted and measured vertical GRFs. ]{
        \includegraphics[width=0.48\linewidth]{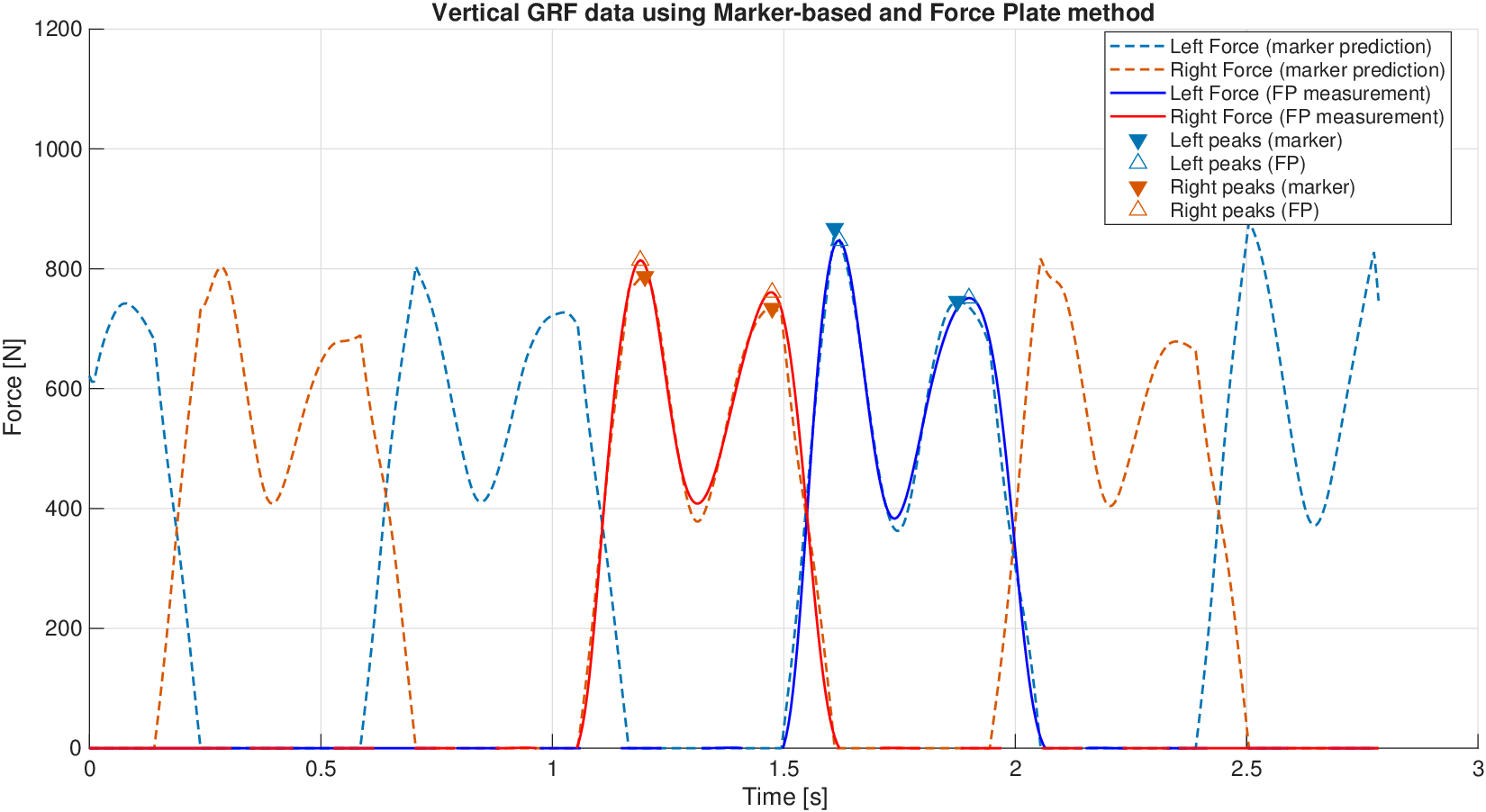}
        \label{fig:individual_GRF_single}
    }
    \hfil
    \subfloat[Summary of R² and peak deviation metrics]{
        \includegraphics[width=0.46\linewidth]{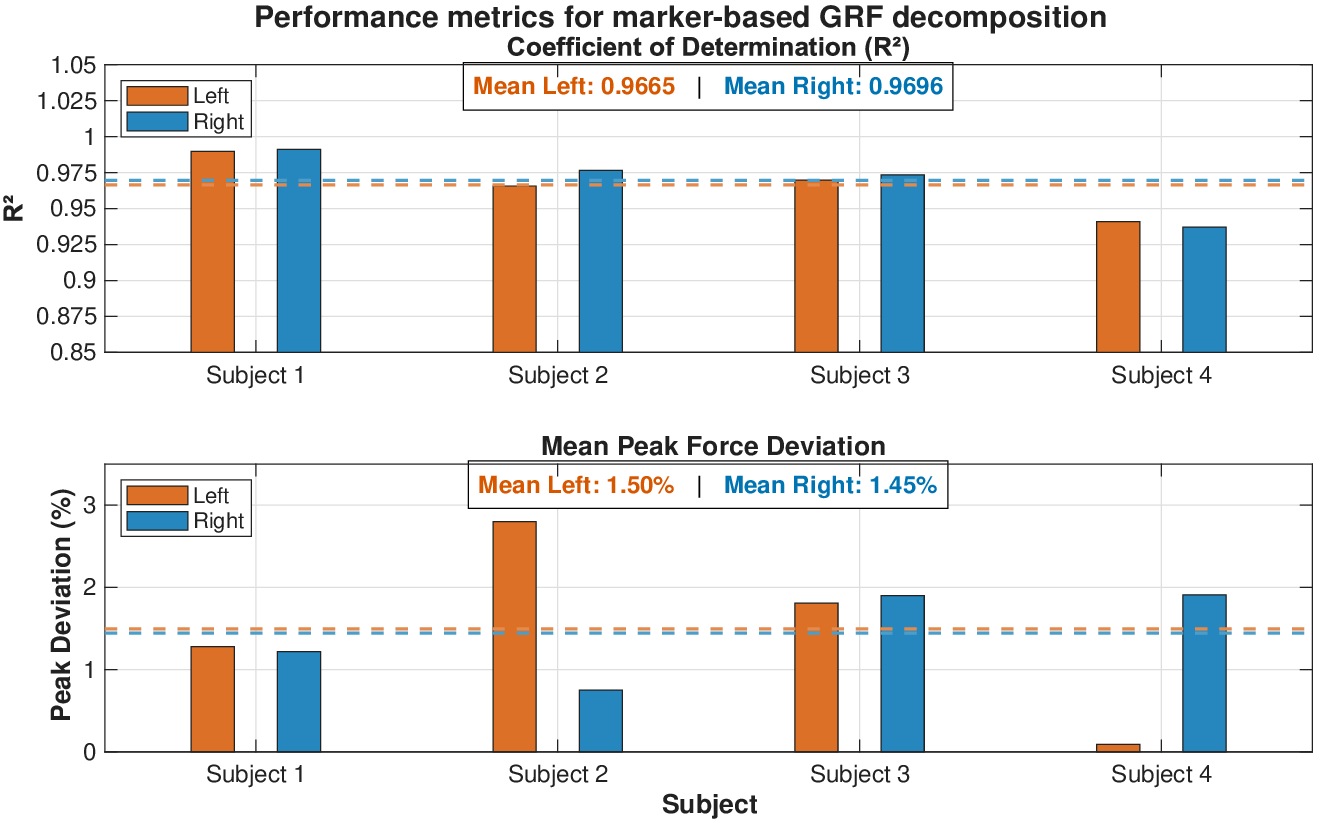}
        \label{fig:individual_GRF_stats}
    }
    \caption{Evaluation of the bilateral GRF decomposition against force plate measurements. (a) Decomposition of the total ground reaction force into discrete left and right foot GRF components based on Eq. \eqref{bilateralForce} for one representative participant. The maxima of both the predicted and force plate measurement signals are also illustrated. (b) Statistical performance metrics of the marker-based GRF decomposition across all subjects, including the coefficient of determination (R²) and mean peak force deviation for both left and right limbs.}
    \label{fig:individual_GRF}
\end{figure*}

By integrating the temporal gait events identified in Fig. \ref{fig:event_detection} with the total estimated GRF trajectories (Fig. \ref{fig:filtered_marker_vs_force_data}), the individual limb contributions were mathematically decoupled. Specifically, Eq. (\ref{bilateralForce}) was applied to resolve the indeterminate distribution of force during the double-support phase, effectively separating the total GRF into discrete left and right kinetic components (Fig. \ref{fig:individual_GRF}). 

To quantitatively evaluate the accuracy of the proposed decomposition, the estimated force components were compared against direct force plate measurements. For the representative subject shown in Fig. \ref{fig:individual_GRF}, the coefficient of determination (R²) between the marker-based estimates and force plate recordings was 0.9409 and 0.9371 for the left and right limbs, respectively. Across all four subjects, the decomposition yielded mean R² values of 0.9665 and 0.9696 for the left and right limbs, indicating strong waveform-level agreement between the predicted and measured vertical GRF. Furthermore, peak force estimation demonstrated excellent accuracy for the representative subject, with deviations of 0.09$\%$ and 1.91$\%$ for the left and right limbs respectively. Averaged across all four subjects, these deviations remained consistently low at 1.50$\%$ and 1.45$\%$ for the left and right limbs respectively, confirming that the magnitude of peak loading, which is of primary biomechanical interest, is reliably captured by the proposed approach. These results collectively demonstrate the validity of the kinematic decomposition framework and support its potential for clinical gait analysis in settings where force plates are unavailable.

The ability to decompose total GRF into bilateral components solely from kinematic data addresses a fundamental limitation of traditional instrumented gait analysis: spatial restriction. While force plates capture only specific footsteps within a calibrated volume, this marker-based decomposition enables the continuous monitoring of limb-specific loading across multiple consecutive stride cycles. This is particularly valuable in clinical scenarios, such as identifying falls and frailty risk, where quantifying load-bearing asymmetry during the weight-transfer phase (double support) is an established early indicator for future falls, yet often difficult to capture repeatedly on fixed force plates and unable to scale for widespread population health screening of gait.

\begin{figure}[h]
    \centering
    \includegraphics[width=1\linewidth]{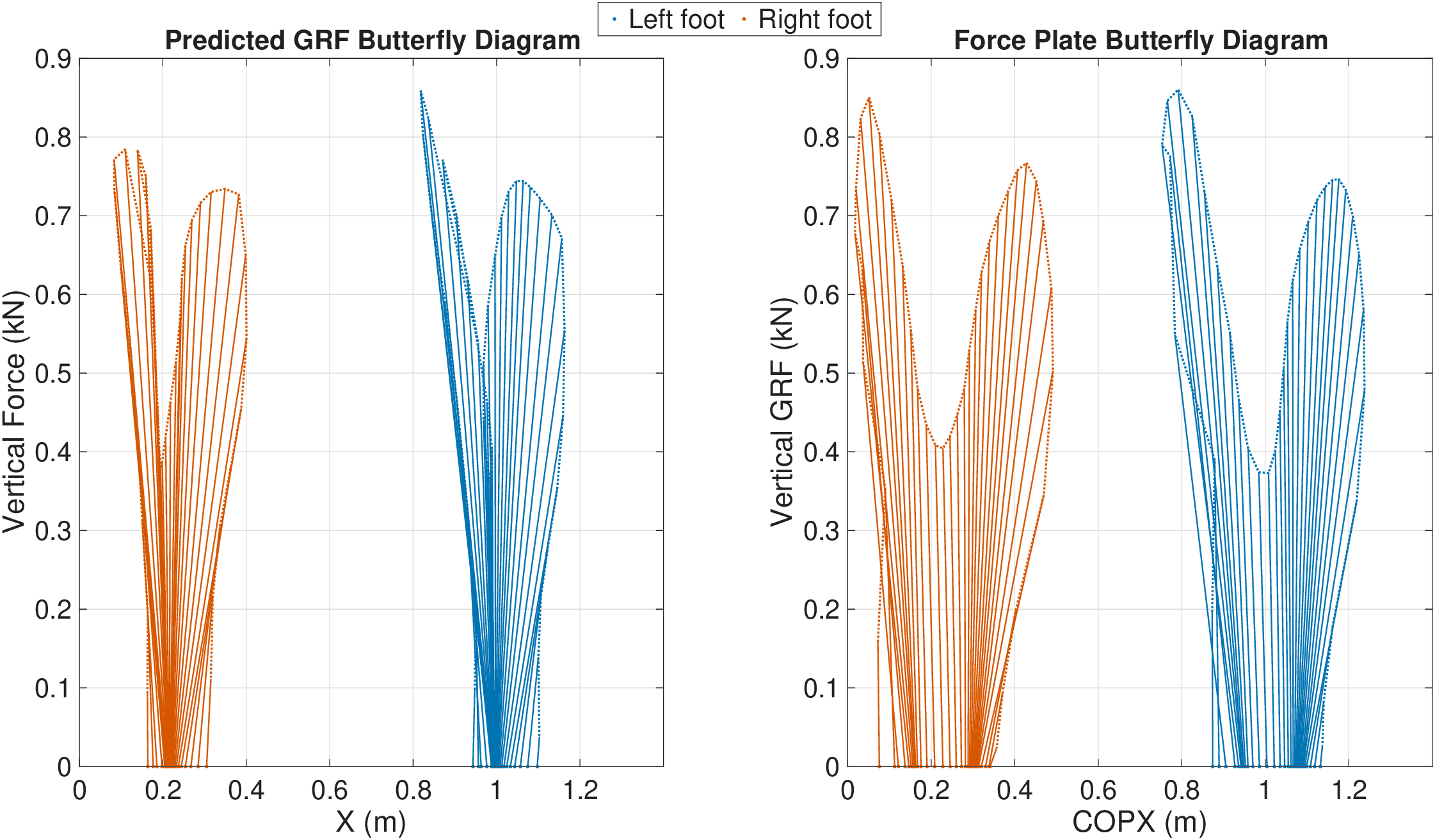}
    \caption{Butterfly diagrams comparing predicted (left) and force-plate (right) GRFs. The derived butterflies successfully reproduce the main shape, peak loading regions, and progression trends observed in the force-plate data}
    \label{fig:butterfly}
\end{figure}

Butterfly diagrams comparing predicted and force-plate ground reaction forces are shown in Fig.~\ref{fig:butterfly}. These diagrams encode key clinical features in a single visualization: loop width reflects weight transfer from heel strike to toe-off, peak height indicates limb loading, and stride-to-stride consistency captures gait repeatability. Left–right asymmetry in peak height, width, or curvature provides a visual indicator of compensatory loading patterns in conditions such as osteoarthritis, hemiplegia, and post-surgical rehabilitation.

The estimated butterfly diagram reproduces the main shape, peak loading regions, and progression trends observed in force-plate data, indicating that the kinematic decomposition captures dominant foot–ground mechanics. Remaining discrepancies in the horizontal spread width arise from using the foot segment centre of mass as a proxy for the true centre of pressure (COP). As the COP shifts from heel to forefoot during stance, this approximation underestimates its forward progression, compressing the predicted butterfly width to some extent. Future work may address this using subject-specific foot models or data-driven COP estimation from plantar pressure insoles.

Although the predicted butterfly diagram cannot fully reproduce the spatial extent of the loading trajectory, it preserves key clinically relevant features. Limb loading magnitude and left–right asymmetry are reliably captured, enabling detection of compensatory strategies and pathological gait. Unlike force plates, which capture only one to two steps per trial, the kinematic approach generates continuous butterfly diagrams across the walking path, allowing stride-by-stride assessment of loop consistency, a direct measure of gait repeatability. This, along with independence from instrumented flooring, extends quantitative gait analysis beyond the laboratory and supports longitudinal monitoring in clinical and resource-limited settings.

Unlike conventional inverse dynamics approaches, which primarily reconstruct joint moments and require assumptions about foot contact models or segmental force transmission, the present method directly estimates GRFs at the whole-body level. By operating at the centre-of-mass rather than recursively at joint level, the approach avoids amplification of kinematic noise across segments and enables continuous GRF estimation across multiple consecutive steps. This distinction reflects a fundamentally different modeling philosophy rather than a variation in implementation.

Because the framework relies only on kinematic input, it naturally complements emerging markerless motion capture systems and high-throughput clinical workflows where force plates are impractical. The ability to estimate and decompose GRFs continuously across multiple strides makes the approach well suited for longitudinal monitoring of fatigue, adaptation, and rehabilitation response. Moreover, the physics-based formulation avoids dataset dependency, supporting scalability across populations, settings, and acquisition platforms.



\section{Conclusion} \label{Conclusion}
In the current work, we demonstrated the viability of CoM and GRF estimation based on only marker position data obtained from the motion capture system. By computing GRFs from whole-body kinematics and decomposing them into its individual components, the proposed framework opens the possibility of measuring GRFs without the need for expensive force plates while retaining access to clinically meaningful kinetic measures. Moreover, force plates are fixed and the infrequent occurrence of step-on events make model-based methods like ours more attractive and practical than current machine learning methods that rely on extensive data. The accuracy of the GRF estimation in gait datasets from pathological populations, i.e. patients with neurological (such as stroke) and musculoskeletal conditions (such as osteoarthritis) remains to be examined. This would be essential to determine its potential for reliable detection and assessment of gait abnormalities across patient populations.

\section*{Ethics Statement}
The experimental procedures outlined in this study were approved by the university’s Institutional Review Board (NTU IRB-2018-04-014) and the study was conducted in accordance with the ethical principles of medical research outlined in the Declaration of Helsinki.

\section*{Acknowledgment}
Any use of generative artificial intelligence was limited to assistance with grammar and syntax corrections, and all content was reviewed and approved by the authors, who take full responsibility for the manuscript.

\section*{APPENDIX}\label{grf_decomposition}
Following standard optimal control literature \cite{Bryson1975}, we seek to determine (sufficiently) \textit{smooth} forces $R_1(t)$ and $R_2(t)$ in the time interval $t\in[t_0\ t_1]$, which 
\begin{subequations}
\begin{itemize}
    \item minimize the following index
        \begin{equation}
             J := \int_{t_0}^{t_1} \left( (\dot R_1(t))^2 +( \dot R_2(t))^2     \right)\,dt
        \end{equation}
    \item subject to
    \begin{itemize}
        \item dynamic constraint (i.e. Newton's law)            \begin{equation}
             R_1(t)+R_2(t)=ma(t)
            \end{equation}
        \item boundary conditions
            \begin{align}\label{eq:BC}
            R_2(t_0) &=0\\  
            R_1(t_1) &=0
            \end{align}
    \end{itemize}
        
\end{itemize}
\end{subequations}

A simple way to solve this problem is to get rid of one of the variables (e.g. $R_2$) via the dynamic constraint (Newton's law), which leads to the minimization the same index
\begin{align}\label{eq:Lagrangian}
J &= \int_{t_0}^{t_1} \Big(\dot R_1^2 
       + \big(m \dot a-\dot R_1 \big)^2 \Big) \, dt \nonumber\\
  &= \int_{t_0}^{t_1} 
       \underbrace{\Big(  2\dot R_1^2  - 2 m \dot R_1  \dot a + m^2\dot a^2 \Big)}_{\text{Lagrangian } \mathcal L} \, dt
\end{align}

The integrand in Eq. \eqref{eq:Lagrangian} is referred to as \textit{Lagrangian}\footnote{In general, the Lagragian $\mathcal L$ would also depend on $R_1$ but, in this specific case, it only depends on $\dot R_1$.}. Via standard variational methods, this minimization problem leads to the Euler-Lagrange equations which, in this case\footnote{In general, Euler-Lagrange equations are written as $\frac{d}{dt}(\partial _{\dot R_1}\mathcal L)-\partial_{R_1}\mathcal L=0$ but, since $\mathcal L$ does not depend on $R_1$, i.e. $\partial_{R_1}\mathcal L=0$, this equation simplifies to $\frac{d}{dt}(\partial _{\dot R_1}\mathcal L)=0$}, take the simple form of
\begin{equation}\nonumber
    \ddot R_1 = \frac{1}{2}m\ddot a
\end{equation}
\REMARK{Symmetrically, a similar result in terms of $R_2$ can be derived if we eliminated $R_1$ instead of $R_2$. Therefore, the minimum rate-of-change model for GRFs leads to:}

\begin{equation}\label{eq:EL}
\boxed{
    \ddot R_1 = \ddot R_2 = \frac{1}{2}m\ddot a
    }
\end{equation}

Direct (double) time-integration, while accounting for the boundary conditions Eq. \eqref{eq:BC}, allows to derive an explicit formulation of the GRFs as:
\begin{subequations} \label{bilateralForce}
\begin{align}
    R_1 (t) &= m\frac{a(t) + a(t_0)}{2} -m\frac{a(t_1)+a(t_0)}{2\, T_{DS}}(t-t_0) \\
    R_2 (t) &= m\frac{a(t) - a(t_0)}{2} +m\frac{a(t_1)+a(t_0)}{2\, T_{DS}}(t-t_0) 
\end{align}
\end{subequations}
where $T_{DS}:=t_1-t_0$ is the duration of the Double-Stance phase (see Fig. \ref{fig:min_GRF}).

\end{document}